\documentclass[letterpaper, 10 pt, conference]{ieeeconf}  % Comment this line out if you need a4paper

\IEEEoverridecommandlockouts                              % This command is only needed if 
\title{\LARGE \bf
The 1st InterAI Workshop: Interactive AI for Human-centered Robotics}

\author{Yuchong Zhang$^{1}$  Elmira Yadollahi$^{2}$  Yong Ma$^{3}$  Di Fu$^{4}$  Iolanda Leite$^{1}$ and Danica Kragic$^{1}$ % <-this % stops a space
\thanks{$^{1}$Yuchong Zhang, Iolanda Leite, and Danica Kragic with Division of Robotics, Perception, and Learning (RPL), KTH Royal Institute of Technology, Stockholm, Sweden
        {\tt\small yuchongz, iolanda, dani@kth.se}}%
\thanks{$^{2}$Elmira Yadollahi with Lancaster University, UK
        {\tt\small e.yadollahi@lancaster.ac.uk}}%
\thanks{$^{3}$Yong Ma with University of Bergen, Norway
        {\tt\small yong.ma@uib.no}}%
\thanks{$^{4}$Di Fu with University of Surrey, UK
        {\tt\small d.fu@surrey.ac.uk}}%
}

\usepackage{url}
\usepackage{hyperref}
\usepackage{balance}

\begin{document}

\maketitle
\thispagestyle{empty}
\pagestyle{empty}

%%%%%%%%%%%%%%%%%%%%%%%%%%%%%%%%%%%%%%%%%%%%%%%%%%%%%%%%%%%%%%%%%%%%%%%%%%%%%%%%
\begin{abstract}

The workshop is affiliated with 33nd IEEE International Conference on Robot and Human Interactive Communication (RO-MAN 2024) August 26~30, 2023 / Pasadena, CA, USA. It is designed as a half-day event, extending over four hours from 9:00 to 12:30 PST time. It accommodates both in-person and virtual attendees (via Zoom), ensuring a flexible participation mode. The agenda is thoughtfully crafted to include a diverse range of sessions: two keynote speeches that promise to provide insightful perspectives, two dedicated paper presentation sessions, an interactive panel discussion to foster dialogue among experts which facilitates deeper dives into specific topics, and a 15-minute coffee break. The workshop website: \href{https://sites.google.com/view/interaiworkshops/home}{link}.

\end{abstract}

%%%%%%%%%%%%%%%%%%%%%%%%%%%%%%%%%%%%%%%%%%%%%%%%%%%%%%%%%%%%%%%%%%%%%%%%%%%%%%%%
\section{Statement of Objectives}

This workshop aims to explore and discuss the advancements and challenges in human-centered interactive artificial intelligence (AI) within the field of human-robot interaction (HRI). It will focus on the integration of AI technologies that enhance human-robot collaboration, ensuring these interactions are intuitive, efficient, and tailored to human needs and behaviors \cite{zhang2024vision,adams2005human,sheridan2016human}.

Based on \cite{schmidt2020interactive}, human-centered interactive AI is defined as an AI that enables interactive exploration and manipulation in real-time and is designed with a clear purpose for human benefit while being transparent about who has control over data and algorithms. This workshop is dedicated to exploring the cutting-edge developments in AI that prioritize interactive, real-time exploration and manipulation, all within the sphere of HRI \cite{obaigbena2024ai,obrenovic2024generative,khandelwal2017bwibots,canal2016real}. We aim to address the design considerations that make AI systems transparent, particularly in terms of data control and algorithmic operations, ensuring that users understand and trust the technology they interact with in the context of human-centered robotics \cite{sreedharan2022explainable,lam2010human,khatib2004human,schaal2007new,zhang2024will,xia2024shaping,zhang2024mind}.

As it’s the first edition, this workshop will serve as a platform for fostering innovation, collaboration, and discussion among the HRI community, driving forward the development of human-centered interactive AI in robotics \cite{he2021challenges,murphy2019introduction,dogmus2014react,rajan2017towards}. Through a series of keynotes, paper presentations, panel discussions, and interactive sessions, we aspire to foster a deep understanding of how human-centered interactive AI can be effectively integrated into HRI systems to create more effective, ethical, and user-friendly interactions.

% \section{Previous Editions}
% This is the first edition of this workshop.

\section{Indented Audience}

\begin{itemize}
    \item The organizers.
    \item Researchers and students in robotics, AI, HCI, and relevant fields.
    \item Industry professionals in robotics and AI development.
    \item Designers, researchers, and developers of HRI systems.
\end{itemize}

\section{List of Speakers}

We have two keynote speakers:

\begin{itemize}
    \item \textbf{Dr. Oya Celiktutan} \footnote{\url{https://nms.kcl.ac.uk/oya.celiktutan/}}. Short bio: Dr. Oya Celiktutan is a Senior Lecturer in Robotics (Associate Professor) at the Department of Engineering, King’s College London, UK, where she leads the Social AI and Robotics Laboratory. Her research, at the intersection of machine learning and human-robot interaction, explores two broad questions through an interdisciplinary lens: how to learn human behavior from multimodal data, and how to transfer this knowledge to robots for learning, action, and interaction. Her work has been supported by EPSRC, The Royal Society, and the EU Horizon, as well as through industrial collaborations. Her team’s research has been recognized with several awards, including the Best Paper Award at IEEE Ro-Man 2022, NVIDIA CCS Best Student Paper Award Runner Up at IEEE FG 2021, First Place Award and Honorable Mention Award at ICCV UDIVA Challenge 2021.
    \item \textbf{Dr. Alessandra Sciutti} \footnote{\url{https://www.iit.it/it/people-details/-/people/alessandra-sciutti}}. Short bio: Alessandra Sciutti is Tenure Track Researcher, head of the CONTACT (COgNiTive Architecture for Collaborative Technologies) Unit of the Italian Institute of Technology (IIT). She received her B.S and M.S. degrees in Bioengineering and the Ph.D. in Humanoid Technologies from the University of Genova in 2010. After two research periods in USA and Japan, in 2018 she has been awarded the ERC Starting Grant wHiSPER (www.whisperproject.eu), focused on the investigation of joint perception between humans and robots. She published more than 80 papers and abstracts in international journals and conferences and participated in the coordination of the CODEFROR European IRSES project (https://www.codefror.eu/). She is currently Associate Editor for several journals, among which the International Journal of Social Robotics, the IEEE Transactions on Cognitive and Developmental Systems and Cognitive System Research. The scientific aim of her research is to investigate the sensory and motor mechanisms underlying mutual understanding in human-human and human-robot interaction.
\end{itemize}

We have four panelists:

\begin{itemize}
    \item Dr. Antonio Andriella \footnote{\url{https://www.antonioandriella.com/}};
    \item Dr. Maria Kyrarini \footnote{\url{sites.google.com/view/mariakyrarini}};
    \item Alva Markelius \footnote{\url{https://alvamarkelius.github.io/}};
    \item Florian Pestoni \footnote{\url{www.linkedin.com/in/florianpestoni}}.
\end{itemize}

\section{List of Topics}
Topics of interest include, but are not limited to:

\begin{itemize}
    \item Human-centered AI for HRI;
    \item Explainability and transparency in AI and robotics;
    \item Design principles for human-centered interactive AI in robotics;
    \item Ethical considerations and societal impact of HRI with AI;
    \item Case studies: successful implementations of interactive AI in HRI;
    \item Enhancing user experience in HRI through interactive AI;
    \item Accessibility in HRI through interactive AI;
    \item Augmenting social and conversational robotics through interactive AI;
    \item Robotics and interactive AI in assistive technologies and healthcare/aging healthcare;
    \item Interactive AI in wearable robotics;
    \item Applications of interactive AI vs generative AI in robotics;
    \item Multimodal social signals processing in HRI through interactive AI
\end{itemize}

\section{Plan to Solicit Participation}

The call for participation will be distributed via mailing lists (i.e., HRI-announcement, CHI-announcements, robotics-worldwide) and social media. To encourage the participation of a multidisciplinary and diverse audience (expected 30 participants), we will also advertise the workshop to mailing lists such as corpora-list (aimed at researchers working on NLP), CVML (aimed at researchers working on computer vision and machine learning), HMC (aimed at researcher working on human-machine communication), the Women in Machine Learning mailing list, and the Women in Robotics Slack channel. A website will be created to provide information about the workshop. Here, prospective participants will find the format and schedule of the workshop, as well as information on the keynotes.

Prospective participants will be invited to submit 2-4 pages of extended abstracts on research related to the central theme and various topics of the workshop. We will encourage the submission of papers describing work-in-progress, preliminary results, or position papers to encourage the discussion and peer review of new ideas. Selected submissions will be invited for oral presentations at the workshop,  providing a platform for authors to share their insights and findings with attendees.

\section{Statement of Inclusion, Diversity and Equity}

This half-day hybrid workshop is designed to maximize participation and accessibility by being available both in-person and via Zoom. This format ensures that both remote and on-site attendees, including those unable to travel to the conference, can actively participate in real-time group discussions, including speakers and attendees from various locations. Scheduled for the morning hours of 9:00 to 12:30 PST, the timing is strategically chosen to accommodate online participants from regions like Europe and Asia as much as possible. To facilitate engagement from remote attendees, a designated member of the organizing committee will oversee the chat, encouraging virtual engagement and assisting with in-person activities on behalf of remote participants as necessary. Additionally, the panel discussion is structured to support online involvement, ensuring an inclusive and interactive experience for all participants.

\section{Workshop Organization}

\begingroup
\renewcommand\labelenumi{(\theenumi)}
\begin{enumerate}
    \item 09:00 - 09:05 Welcome and introduction to the workshop
    \item 09:05 - 09:30 Keynote (20 + 5 Q/A): Keynote Presentation 1 (20 + 5 Q\&A) with Dr. Alessandra Sciutti
    \item 09:30 - 10:15 Paper Presentations 1 (5 +2 Q\&A)
    \item 10:15 - 10:30 Coffee break
    \item 10:30 - 10:45 Paper Presentations 2 (5 +2 Q\&A)
    \item 10:45 - 11:10 Keynote Presentation 2 (20 + 5 Q/A) with Dr. Oya Celiktutan
    \item 11:10 - 11:25 Duckietown + Awards Announcement
    \item 11:25 - 12:25 Panel Discussion
    \item 12:25 - 12:30 Concluding Remarks
\end{enumerate}
\endgroup

\section{Sponsors}
We are excited to announce that our workshop is proudly sponsored by Duckietown (\url{https://duckietown.com/}), which will be presenting three Best Paper Awards, and InOrbit (\url{https://www.inorbit.ai/}), which will sponsor the Best Oral Presentation Award. We extend our gratitude to both sponsors for their generous support in recognizing outstanding contributions.

% \section{Expected Devices}
% We anticipate a meeting room of adequate size furnished with essential items, including desks, tables, chairs, and whiteboards. Additionally, the room should be equipped with comprehensive video conferencing equipment to support seamless virtual connection and discussion.

\section{Organizers}

\textbf{Yuchong Zhang} is currently a postdoctoral fellow at KTH Royal Institute of Technology in Sweden. He obtained his Ph.D. degree from Chalmers University of Technology in 2023 and MSc. degree from Nanyang Technological University in Singapore in 2017. His current research interest include human-robot interaction with human-centered design, interactive AI, VR/AR/MR, human perception, and affective computing. He was previously working as a Marie Curie Early Stage Researcher (MSCA Horizon 2020 ITN) from 2018 to 2022 specializing in the TOMOCON project (\url{https://www.tomocon.eu/}). He actively serves as an AC/AE at several venues, such as ACM CHI, ACM CSCW, MuC, CHIPlay, HCII, ICSR, etc.
% \begin{itemize}
%     \item Affiliation: KTH Royal Institute of Technology
%     \item Address: Room 308, Lindstedtsvägen 24, 114 28 Stockholm, Sweden
%     \item Phone: +46-701493091\\
%     \item Email Address:  yuchongz@kth.se \\
%     \item Website: https://sites.google.com/view/yuchongzhang
% \end{itemize}

\textbf{Elmira Yadollahi} is an assistant professor (UK lecturer) in computer science at Lancaster University in the UK. Previously, she was a Postdoctoral Fellow at the Division of Robotics, Perception, and Learning (RPL) at KTH Royal Institute of Technology in Sweden since October 2021. She obtained her PhD in 2021 in Robotics and Computer Science from École Polytechnique Fédérale de Lausanne (EPFL), Switzerland and Instituto Superior Técnico, Portugal. Her research tackles a range of topics on explainability in human-robot interaction, educational child-robot interaction, robot cognitive development, and assistive robotics. She is an associate editor of the International Journal of Child-Computer Interaction (IJCCI) and has served as a guest editor at the Interaction Studies Journal. She served as co-chair of the Research and Design Challenge Track at the ACM Interaction Design and Children Conference (IDC) in 2023 and 2024 and as workshop chair at Human-Agent Interaction (HAI) in 2023.  She has co-organized several workshops at HRI, ICSR, and IDC conferences.
% \begin{itemize}
%     \item Affiliation: KTH Royal Institute of Technology
%     \item Phone: +46 764533420
%     \item Email Address:  elmiray@kth.se
%     \item Website: https://elmirayadollahi.com/
% \end{itemize}

\textbf{Yong Ma} currently works as a postdoctoral researcher at the University of Bergen in Norway. He obtained his Ph.D. degree from LMU of Munich. His research interests include speech signal processing, voice interface, human-machine interaction, machine learning, dementia diagnosis, and healthcare.
% \begin{itemize}
%     \item Affiliation: University of Bergen 
%     \item Phone: +49 15207660245
%     \item Email Address: yong.ma@uib.no 
%     \item Website: https://www.uib.no/en/persons/Yong.Ma
% \end{itemize}

\textbf{Di Fu} is an assistant professor (UK lecturer) at the Department of Psychology, University of Surrey. Her research group focuses on crossmodal learning and human-robot social interaction. Before that, she worked as a postdoctoral research associate at the Department of Informatics, University of Hamburg. She completed her doctoral training in human-robot interaction with Prof. Stefan Wermter from 2017 to 2020 and cognitive neuroscience at the Institute of Psychology, Chinese Academy of Sciences (CAS) with Prof. Xun Liu from 2014 to 2020. She had been honored as an outstanding graduate of CAS and an outstanding doctoral graduate of Beijing. She has been awarded the Kavli Summer Institute in Cognitive Neuroscience fellowship, the International Postdoctoral Exchange fellowship, the CAS-DAAD joint doctoral student fellowship, and the Chinese National Academic Scholarship. Her work has been published in the International Journal of Social Robotics, Public Administration Review, IEEE IROS, ACM/IEEE HRI, IEEE RO-MAN, IEEE IJCNN, etc. She also serves as a committee member of the Chinese Association for Psychological \& Brain Sciences and the Chinese German Association for Biology and Medicine.
% \begin{itemize}
%     \item Affiliation: University of Surrey
%     \item Phone: +49 15165043909
%     \item Email Address:  d.fu@surrey.ac.uk 
%     \item Website: https://www.difu-academic.org/ 
% \end{itemize}

\textbf{Iolanda Leite} is an Associate Professor at the School of Electrical Engineering and Computer Science at KTH Royal Institute of Technology, Sweden. She holds a PhD in Information Systems and Computer Engineering from IST, University of Lisbon. Prior to joining KTH, she was a Postdoctoral Associate at the Yale Social Robotics Lab and an Associate Research Scientist at Disney Research Pittsburgh. Her research goal is to develop robots that can perceive, learn from, and respond appropriately to people in real-world situations.
% Affiliation: KTH Royal Institute of Technology
% Phone: +46 72 566 09 98
% Email Address: iolanda@kth.se
% Website: https://iolandaleite.com/ 

\textbf{Danica Kragic} is a Professor at the School of Computer Science and Communication at the Royal Institute of Technology, KTH. She received MSc in Mechanical Engineering from the Technical University of Rijeka, Croatia in 1995 and PhD in Computer Science from KTH in 2001. She has been a visiting researcher at Columbia University, Johns Hopkins University and INRIA Rennes. She is the Director of the Centre for Autonomous Systems. Danica received the 2007 IEEE Robotics and Automation Society Early Academic Career Award. She is a member of the Royal Swedish Academy of Sciences, Royal Swedish Academy of Engineering Sciences and Young Academy of Sweden. She holds a Honorary Doctorate from the Lappeenranta University of Technology. She chaired IEEE RAS Technical Committee on Computer and Robot Vision and served as an IEEE RAS AdCom member. Her research is in the area of robotics, computer vision and machine learning. In 2012, she received an ERC Starting Grant. Her research is supported by the EU, Knut and Alice Wallenberg Foundation, Swedish Foundation for Strategic Research and Swedish Research Council. She is an IEEE Fellow.
% Affiliation: KTH Royal Institute of Technology
% Phone: +46 87 906 729
% Email Address: dani@kth.se  
% Website: https://www.csc.kth.se/~danik/ 

\section*{ACKNOWLEDGMENT}
At KTH, this workshop is partially funded by grants from the Swedish Foundation for Strategic Rese
arch (SSF FFL18-0199 and FUS21-0067), the S-FACTOR project from NordForsk, Digital Futures, the Vinnova Competence Center for Trustworthy Edge Computing Systems and Applications, and the HORIZON-CL4-2021-HUMAN-01 ELSA project.

% \begin{thebibliography}{99}

% \end{thebibliography}

\balance
\bibliographystyle{IEEEtran}
\bibliography{Reference}

\end{document}